\documentclass[10pt,twocolumn,letterpaper]{article}

\usepackage[pagenumbers]{cvpr} 
\usepackage{makecell}
\usepackage{dblfloatfix}
\usepackage{placeins}

\usepackage[dvipsnames, table]{xcolor}
\usepackage{graphicx}
\usepackage{multirow}
\usepackage{tabularx}
\usepackage{wrapfig}
\usepackage[acronym,toc]{glossaries}
\usepackage[flushleft]{threeparttable}

\definecolor{cvprblue}{rgb}{0.21,0.49,0.74}
\usepackage[pagebackref,breaklinks,colorlinks,citecolor=cvprblue]{hyperref}

\def\confName{CVPR}
\def\confYear{2024}

\title{Team LEYA in 10th ABAW Competition: Multimodal Ambivalence/Hesitancy Recognition Approach}

\author{Elena Ryumina\\
St. Petersburg Federal Research Center \\ of the Russian Academy of Sciences\\
St. Petersburg, Russia\\
{\tt\small ryumina.e@iias.spb.su}
\and
Alexandr Axyonov\\
St. Petersburg Federal Research Center \\ of the Russian Academy of Sciences\\
St. Petersburg, Russia\\
{\tt\small axyonov.a@iias.spb.su}
\and
Dmitry Sysoev \\
HSE University\\
St. Petersburg, Russia\\
{\tt\small dsysoev@hse.ru}
\and
Timur Abdulkadirov \\
HSE University\\
St. Petersburg, Russia\\
{\tt\small tnabdulkadirov@edu.hse.ru}
\and
Kirill Almetov \\
HSE University\\
St. Petersburg, Russia\\
{\tt\small koalmetov@edu.hse.ru}
\and
Yulia Morozova \\
HSE University\\
St. Petersburg, Russia\\
{\tt\small yuvmorozova\_1@edu.hse.ru}
\and
Dmitry Ryumin\\
HSE University\\
St. Petersburg Federal Research Center \\ of the Russian Academy of Sciences\\
St. Petersburg, Russia\\
{\tt\small daryumin@hse.ru}
}

\begin{document}
\newacronym{FCL}{FCL}{Fully Connected Layer}
\newacronym{FPS}{FPS}{Frame Per Second}
\newacronym{LSTM}{LSTM}{Long Short-Term Memory}
\newacronym{BiLSTM}{BiLSTM}{Bidirectional Long Short-Term Memory}
\newacronym{VAD}{VAD}{Voice Activity Detection}
\newacronym{SOTA}{SOTA}{State-of-the-Art}
\newacronym{BAH}{BAH}{Behavioural Ambivalence/Hesitancy}
\newacronym{ViT}{ViT}{Vision Transformer}
\newacronym{CLIP}{CLIP}{Contrastive Language-Image Pretraining}
\newacronym{LLM}{LLM}{Large Language Model}
\newacronym{MLP}{MLP}{Multilayer Perceptron}
\newacronym{TF-IDF}{TF-IDF}{Term Frequency-Inverse Document Frequency}
\newacronym{A/H}{A/H}{Ambivalence/Hesitancy}
\newacronym{ABAW}{ABAW}{Affective \& Behavior Analysis in-the-Wild}
\newacronym{MF1}{MF1}{Macro F1-score}
\newacronym{BERT}{BERT}{Bidirectional Encoder Representations from Transformers}
\newacronym{GELU}{GELU}{Gaussian Error Linear Unit}
\newacronym{GBM}{GBM}{Gradient Boosting Machines}
\newacronym{SGD}{SGD}{Stochastic Gradient Descent}
\newacronym{VideoMAE}{VideoMAE}{Video Masked Autoencoder}
\newacronym{SA}{SA}{Self-Attention}
\newacronym{LR}{LR}{Learning Rate}
\newacronym{LS}{LS}{Label Smoothing}

\maketitle
\glsresetall
\begin{abstract}
Ambivalence/hesitancy recognition in unconstrained videos is a challenging problem due to the subtle, multimodal, and context-dependent nature of this behavioral state. In this paper, a multimodal approach for video-level ambivalence/hesitancy recognition is presented for the 10th ABAW Competition. The proposed approach integrates four complementary modalities: scene, face, audio, and text. Scene dynamics are captured with a VideoMAE-based model, facial information is encoded through emotional frame-level embeddings aggregated by statistical pooling, acoustic representations are extracted with EmotionWav2Vec2.0 and processed by a Mamba-based temporal encoder, and linguistic cues are modeled using fine-tuned transformer-based text models. The resulting unimodal embeddings are further combined using multimodal fusion models, including prototype-augmented variants. Experiments on the BAH corpus demonstrate clear gains of multimodal fusion over all unimodal baselines. The best unimodal configuration achieved an average MF1 of 70.02\%, whereas the best multimodal fusion model reached 83.25\%. The highest final test performance, 71.43\%, was obtained by an ensemble of five prototype-augmented fusion models. The obtained results highlight the importance of complementary multimodal cues and robust fusion strategies for ambivalence/hesitancy recognition. The source code is publicly available\footnote{\url{https://github.com/LEYA-HSE/ABAW10-BAH}}.
\end{abstract}

\glsresetall

\section{Introduction}
\label{sec:intro}

Affective computing aims to endow intelligent systems with the ability to perceive, model, and interpret human affect from signals such as facial behavior~\cite{sajjad2023comprehensive}, speech~\cite{george2024review}, language~\cite{deng2021survey}, and body motion~\cite{leong2023facial}. This capability is important for human-computer interaction, digital health, education, and assistive technologies, where decisions often depend on subtle and context-dependent behavioral cues~\cite{poria2017review}. Within this area, the \gls{A/H} Video Recognition Challenge of the 10th Workshop and Competition on \gls{ABAW} focuses on a particularly difficult binary task: given a video, the goal is to predict whether it contains \gls{A/H} or not at the video level~\cite{gonzalezbah}.

\gls{A/H} recognition is important because these states are strongly linked to decision uncertainty, resistance, and fluctuating motivation during behavior change. In digital behavioral health interventions, such signals can help identify whether a person is ready to change, struggling with conflicting intentions, or at risk of disengagement~\cite{bijkerk2024engagement}. Unlike basic emotions (such as happiness, surprise), \gls{A/H} is subtle and often manifests through inconsistencies across modalities, for example, between what a person says, how they say it, and how they look while speaking. This makes the task inherently multimodal and particularly challenging~\cite{gonzalezbah,poria2017review}.

\citet{gonzalezbah} established baselines using visual facial information, audio, and speech transcripts, and showed that text was the best unimodal cue, while multimodal fusion could yield additional gains but required specialized designs to capture cross-modal inconsistencies characteristic of \gls{A/H}. Similarly,~\citet{Savchenko_2025_CVPR} explored face, audio, and text modalities, with the best validation result achieved by combining textual and facial features, whereas~\citet{Hallmen_2025_CVPR} used text, vision, and audio, and their best final \gls{BAH} submission was based on trimodal fusion. Therefore, multimodal modeling remains a promising direction for \gls{A/H} recognition, especially when modality interactions are explicitly taken into account and fusion preserves modality-specific cues uunder uncertainty or partially inconsistent multimodal evidence~\cite{fang2025emoe,yang2025uncertain}.

This work proposes a multimodal approach that integrates audio, text, face, and scene information for video-level \gls{A/H} recognition. First, a dedicated unimodal model is trained for each modality to learn compact modality-specific representations. 
The resulting modality embeddings are then projected into a shared latent space and fused by a Transformer-based multimodal module operating on modality tokens and augmented with a prototype-based classification objective. In this way, inter-modality dependencies are modeled directly at the video level, while the strengths of the specialized unimodal encoders are preserved for the final \gls{A/H} prediction.

\section{Related Work}
\label{sec:rw}

~\citet{gonzalezbah} presented a broad benchmark suite including supervised unimodal, bimodal, and trimodal models based on facial video, audio, and speech transcripts, as well as zero-shot and personalization experiments. To study multimodal learning, they compared several fusion strategies, including concatenation-based fusion, co-attention, transformer-based fusion, and cross-attention fusion. Their results showed that text is the strongest unimodal cue, while effective \gls{A/H} recognition requires specialized multimodal and temporal modeling to capture subtle inconsistencies across modalities.

Among the approaches proposed for the first public \gls{A/H} challenge,~\citet{Hallmen_2025_CVPR} used a trimodal architecture combining text, vision, and audio. Their framework extracts visual features with a \gls{ViT}~\cite{caron2021emerging}, audio representations with Wav2Vec~2.0~\cite{baevski2020wav2vec}, and transcript embeddings with the \gls{BERT}~\cite{devlin2019bert} text encoder, applies temporal modeling to the visual and audio streams with \glspl{LSTM}~\cite{hochreiter1997long}, and fuses the resulting modality-specific representations with an \gls{MLP}~\cite{rumelhart1986learning}. Their analysis showed that text is the most informative modality. The best final BAH submission was obtained using trimodal fusion, which is consistent with recent studies~\cite{fang2025emoe,li2025infobridge} indicating that preserving complementary modality-specific information during fusion improves multimodal prediction.

In contrast,~\citet{Savchenko_2025_CVPR} used a lighter pipeline based on facial, acoustic, and textual features extracted with EmotiEffLib~\cite{savchenko2024emotieffnet}, Wav2Vec~2.0~\cite{baevski2020wav2vec}, and RoBERTa-based~\cite{liu2019roberta} text representations. Audio and text features were aligned to the visual timeline by interpolation, and fusion was implemented either through early concatenation followed by a feed-forward classifier or through blending of unimodal predictions. In addition, a video-level logistic regression model over pooled text features was used as a filtering step. Their experiments again showed that text is the strongest unimodal modality, while the best validation result was achieved by combining text and facial information.

In summary, prior work shows that text is the strongest unimodal cue for \gls{A/H}, while multimodal fusion remains beneficial, especially when it preserves complementary information across modalities and accounts for uncertainty in multimodal evidence. Unlike previous approaches, which mainly rely on face, audio, and text with relatively simple fusion schemes, our approach additionally incorporates scene information and uses a Transformer-based fusion module over modality-specific representations learned by dedicated unimodal models, further regularized by a prototype-based classification objective.
\section{Proposed Approach}
\label{sec:method}

\begin{figure*}[!t]
  \centering
   \includegraphics[width=1.0\linewidth]{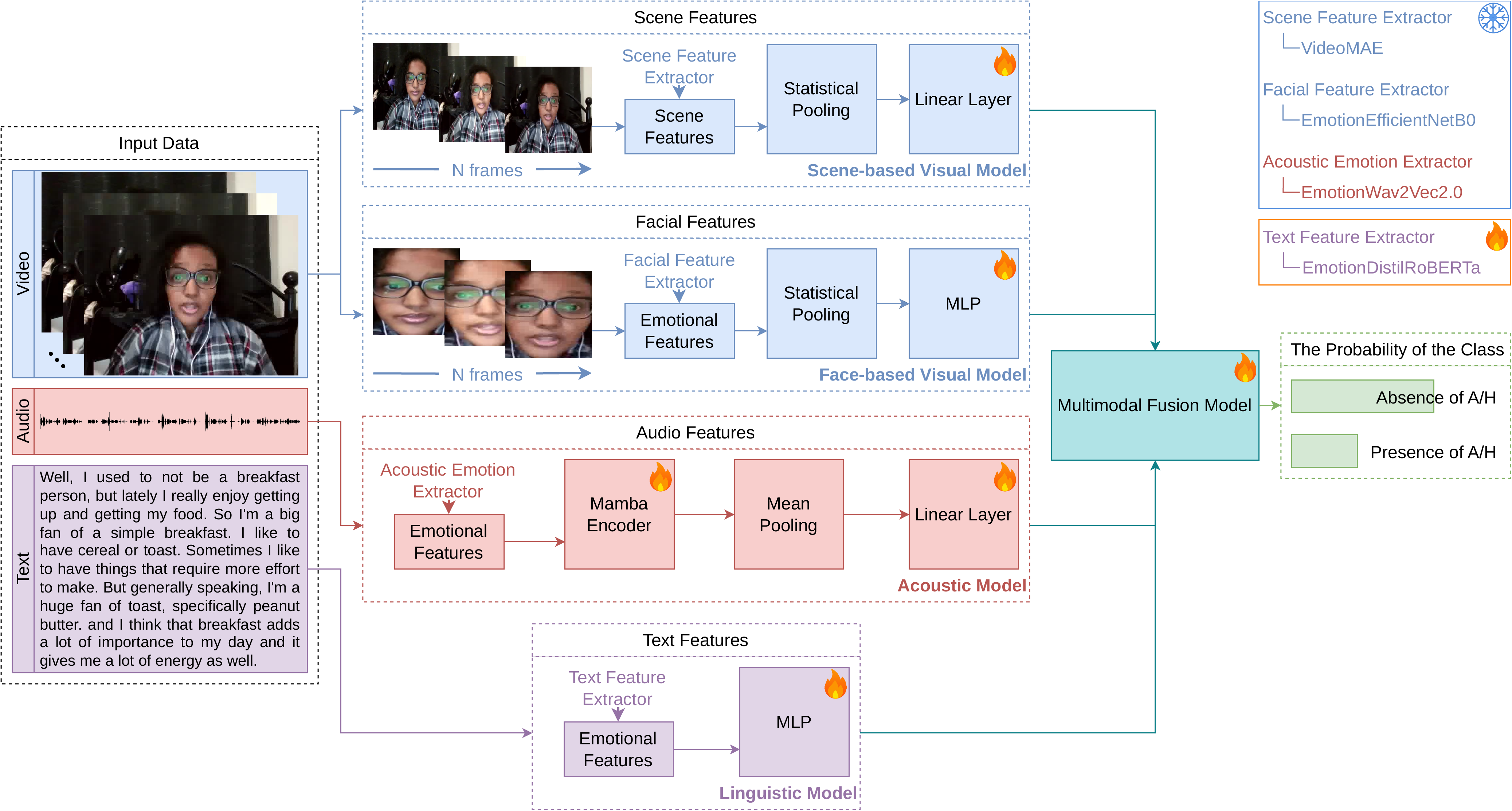}
   \caption{Pipeline of the proposed multimodal approach.}
   \label{fig:pipeline}
\end{figure*}

The overall pipeline of the proposed approach is shown in Figure~\ref{fig:pipeline}. Our approach addresses multimodal \gls{A/H} recognition by extracting complementary information from several synchronized modalities. The obtained representations are then aggregated and fused to produce the final prediction. The following subsections describe the main components of the proposed approach.

\subsection{Scene-based Visual Model}
\label{subsec:svm}

To analyze behavioral dynamics and detect uncertainty, we employ the \gls{VideoMAE} architecture~\cite{tong2022videomae} based on the \gls{ViT}~\cite{vit2020} framework and pre-trained on the Kinetics-400 corpus~\cite{kay2017kinetics}. For each video, $T_v=16$ frames are uniformly sampled, resized to $224 \times 224$, and normalized using ImageNet~\cite{deng2009imagenet} statistics. The input video clip is processed using tubelet embedding, where it is partitioned into non-overlapping spatio-temporal patches of size $2 \times 16 \times 16$. These tubelets are projected into a $D=768$ latent space and combined with learnable positional embeddings.



The resulting tokens are processed by a Transformer encoder with spatio-temporal self-attention to model spatio-temporal dependencies. A compact scene embedding $h_s$ is obtained by applying global average pooling to the output tokens:
\begin{equation}
h_s = \frac{1}{N} \sum_{i=1}^{N} z_i,
\end{equation}
where $N$ is the number of output tokens and $z_i$ denotes the representation of the $i$-th token.

\subsection{Face-based Visual Model}
\label{subsec:fvm}

For each video, frames are uniformly sampled, and a YOLO-based face detector\footnote{\url{https://github.com/lindevs/yolov8-face}} is applied to each sampled frame. When multiple faces are detected, the largest bounding box is selected. If no face is detected, the full frame is used as a fallback crop.

The cropped face is resized to $224\times224$ and normalized with ImageNet~\cite{deng2009imagenet} statistics, then passed to 
an EfficientNetB0~\cite{tan2019efficientnet} extractor fine-tuned on the AffectNet+ corpus~\cite{fard2025affectnetplus}, hereafter referred to as EmotionEfficientNetB0. The extractor produces one emotional embedding vector per sampled frame. No extra embedding normalization is applied.

For each video, frame-level embeddings $\{e_f\}_{f=1}^{F}$ are aggregated using statistical pooling:
\begin{equation}
\mu = \frac{1}{F}\sum_{f=1}^{F} e_f, \qquad
\sigma = \sqrt{\frac{1}{F}\sum_{f=1}^{F}(e_f - \mu)^2}.
\end{equation}

\begin{figure}[!t]
  \centering
   \includegraphics[width=1.0\linewidth]{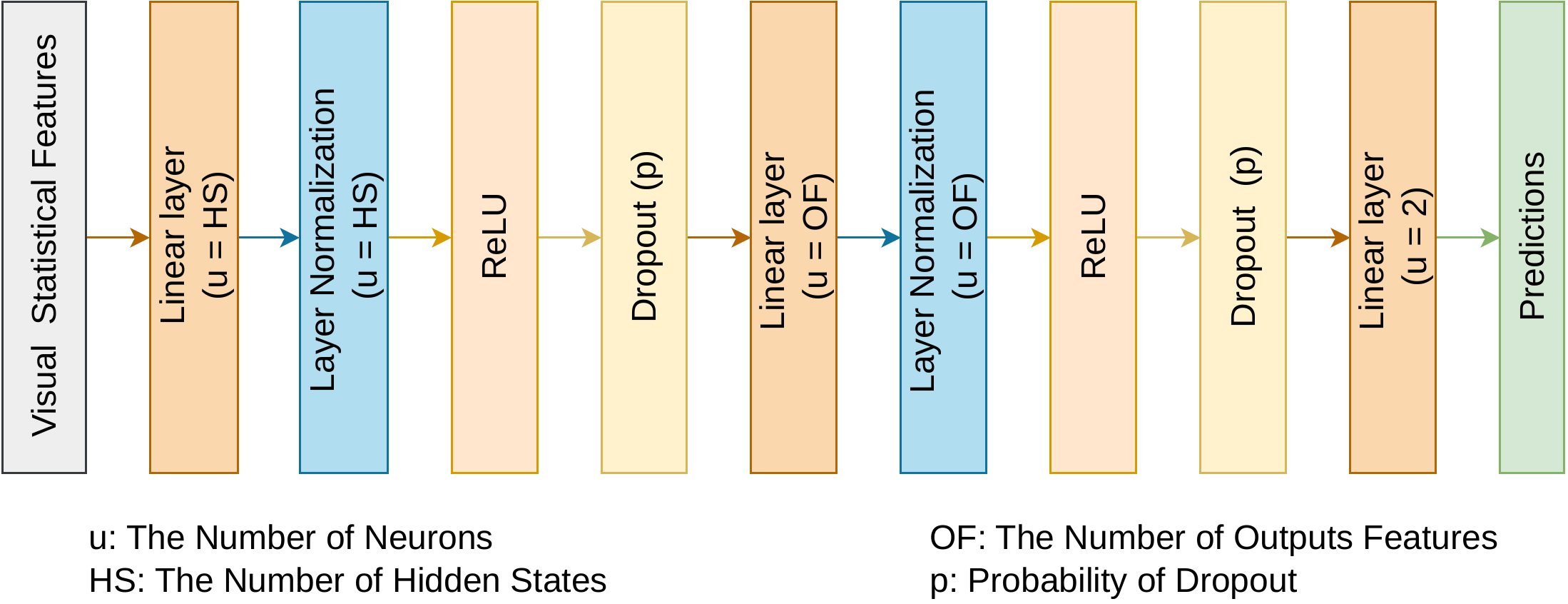}
   \caption{Visual \gls{MLP} architecture.}
   \label{fig:visual_mlp}
\end{figure}

The final face representation is formed by the concatenation $[\mu; \sigma]$. The resulting statistical representation is then used as input to an \gls{MLP}, as shown in Figure~\ref{fig:visual_mlp}.

\subsection{Acoustic Model}
\label{subsec:am}

For the audio modality, the audio track is extracted from each video and resampled to 16 kHz. A pre-trained Wav2Vec2.0 model~\cite{Wagner_2023}\footnote{\url{https://huggingface.co/audeering/wav2vec2-large-robust-12-ft-emotion-msp-dim}}, fine-tuned on the MSP-Podcast corpus~\cite{lotfian2017building}, is used to extract acoustic emotion features. For simplicity, this encoder is referred to as EmotionWav2Vec2.0.

As a result, each audio segment is represented as a sequence of embeddings of size $T_a \times 1024$, where $T_a$ denotes the number of temporal steps and $1024$ is the feature dimension. The value of $T_a$ depends on the duration of the input video.

The extracted acoustic representations are then processed by a sequential encoder to model temporal dependencies in the speech signal. In the main configuration, the extracted acoustic representations are processed by a Mamba encoder~\cite{gu2024mamba}, followed by mean pooling over the temporal dimension to obtain a compact acoustic embedding:
\begin{equation}
a = \frac{1}{T_a}\sum_{t=1}^{T_a} s_t,
\end{equation}
where $s_t$ denotes the representation at the $t$-th temporal step and $a$ is the pooled acoustic embedding. The resulting embedding is then passed through a linear layer to produce the final prediction. In addition to the Mamba-based~\cite{gu2024mamba} variant, a Transformer-based~\cite{vaswani2017attention} encoder was also evaluated as an alternative sequential architecture.

\subsection{Linguistic Model}
\label{subsec:lm}

For the linguistic modality, all audio recordings are transcribed into text. These transcriptions were automatically extracted by the corpus authors~\cite{gonzalezbah}.

Several text-based modeling strategies are considered. A classical \gls{TF-IDF}~\cite{saltonVectorSpaceModel1975} representation is first used to encode the relative importance of words and phrases in each transcription. The resulting features are combined with conventional classifiers, including logistic regression and \gls{GBM} models such as LightGBM~\cite{keLightGBMHighlyEfficient2017} and CatBoost~\cite{prokhorenkova2018catboost}.

Pre-trained transformer-based language models are also used to obtain contextualized text representations. EmotionDistilRoBERTa\footnote{\url{https://huggingface.co/j-hartmann/emotion-english-distilroberta-base/}},   allMiniLM\footnote{\url{https://huggingface.co/sentence-transformers/all-MiniLM-L6-v2}}, EmotionTextClassifier\footnote{\url{https://huggingface.co/michelleli99/emotion_text_classifier}}, and DeBERTa~\cite{he2021deberta} are used to encode the transcriptions into dense embeddings. Both the \texttt{[CLS]} token representation and mean pooling over token embeddings are explored, followed by a \gls{GBM}-based classifier.

In the main configuration, EmotionDistilRoBERTa is directly fine-tuned for \gls{A/H} recognition. Its output representation is passed through an \gls{MLP}-based classification head to produce the final prediction. Other fine-tuned transformer variants, including EmotionTextClassifier, are also evaluated.

\subsection{Modality Fusion Model}
\label{subsec:mfm}

A two-stage strategy is adopted. First, unimodal models for scene, face, audio, and text are trained independently on the target corpus. Then, for each video, one fixed-dimensional embedding per modality is extracted from the corresponding unimodal branch and used as input to the fusion model.

Let $M$ be the number of modalities, $m \in \{1,\dots,M\}$ the modality index, and $x_m \in \mathbb{R}^{d_m}$ the input embedding of modality $m$, where $d_m$ denotes its original dimensionality. Each modality is projected into a shared latent space $\mathbb{R}^{d}$:
\begin{equation}
u_m = \phi_m(x_m), \qquad u_m \in \mathbb{R}^{d},
\end{equation}
where $\phi_m$ is a modality-specific projector composed of a linear layer, layer normalization, \gls{GELU}, and dropout. The resulting modality tokens are stacked into a matrix:
\begin{equation}
U = [u_1;\dots;u_M] \in \mathbb{R}^{M \times d}.
\end{equation}

If one or more modality embeddings are unavailable for a given sample, a binary modality mask is provided to the fusion encoder so that the corresponding tokens are masked out during self-attention. A learnable modality embedding matrix $E_{\mathrm{mod}} \in \mathbb{R}^{M \times d}$ is added to the token sequence:
\begin{equation}
Z^{(0)} = U + E_{\mathrm{mod}},
\end{equation}
where $Z^{(0)}$ is the input token sequence to the Transformer encoder.

The token sequence is processed by a stack of Transformer encoder layers:
\begin{equation}
Z^{(l+1)} = T^{(l)}(Z^{(l)}), \qquad l = 0,\dots,L-1,
\end{equation}
where $T^{(l)}(\cdot)$ denotes the $l$-th Transformer encoder block and $L$ is the number of layers. 
The final fused representation is obtained by masked mean pooling over the output modality tokens:
\begin{equation}
z_{\mathrm{fused}} =
\frac{\sum_{m=1}^{M} \mu_m z_m^{(L)}}{\sum_{m=1}^{M} \mu_m},
\end{equation}
where $z_m^{(L)}$ denotes the output representation of modality $m$ at the last encoder layer and $\mu_m \in \{0,1\}$ indicates whether modality $m$ is available for the given sample.

For the prototype-augmented variant, the fused representation is compared with a set of learnable class-specific prototypes. Let $P_c = \{p_{c,k}\}_{k=1}^{K}$ be the prototype set for class $c$, where $K$ is the number of prototypes per class. The similarity score for class $c$ is computed as:
\begin{equation}
\hat{y}^{\mathrm{proto}}_c =
\log \sum_{k=1}^{K}
\exp\left(\frac{\tilde{z}_{\mathrm{fused}}^\top \tilde{p}_{c,k}}{\tau}\right),
\end{equation}
where $\tilde{z}_{\mathrm{fused}}$ and $\tilde{p}_{c,k}$ denote the $\ell_2$-normalized fused representation and class prototypes, respectively, and $\tau$ is a temperature parameter. In the implemented model, the prototype head does not directly produce the final multimodal prediction; instead, it contributes an auxiliary loss term during training, while the final output logits are produced by the main linear classifier.

Accordingly, the overall training objective is defined as:
\begin{equation}
\mathcal{L}
=
\mathcal{L}_{\mathrm{cls}}
+
\lambda_{\mathrm{proto}} \mathcal{L}_{\mathrm{proto}}
+
\lambda_{\mathrm{div}} \mathcal{L}_{\mathrm{div}},
\end{equation}
where $\mathcal{L}_{\mathrm{cls}}$ is the main classification loss computed from the output logits of the linear classifier, $\mathcal{L}_{\mathrm{proto}}$ is the auxiliary classification loss computed from the prototype-based logits, and $\mathcal{L}_{\mathrm{div}}$ is the prototype diversity regularization term.

\section{Experiments}
\label{sec:exps}

\subsection{Research Corpus}
\label{subsec:rs}

The \gls{BAH} corpus is the core research corpus for the 10th \gls{ABAW} \gls{A/H} challenge. It was collected to support multimodal recognition of ambivalence and hesitancy in videos recorded in a realistic digital behavior change scenario~\cite{gonzalezbah}. Participants answered a predefined set of questions designed to elicit neutral, positive, negative, willing, resistant, ambivalent, and hesitant responses. The data were collected through an online platform with an avatar-guided interaction setup in order to mimic real-world personalized behavioral interventions~\cite{gonzalezbah}.

The full corpus contains $1{,}427$ videos from $300$ participants, totaling $10.60$ hours of recordings. It includes video-level and frame-level expert annotations, timestamps of \gls{A/H} segments, cropped and aligned faces, speech transcripts with timestamps, and participant metadata~\cite{gonzalezbah}. Since ambivalence and hesitancy are difficult to separate reliably in practice, the task is formulated as a binary classification problem indicating the presence or absence of \gls{A/H}~\cite{gonzalezbah}. For the 10th \gls{ABAW} \gls{A/H} challenge, the corpus is partitioned participant-wise into training, validation, public test, and private test subsets. Video-level prediction is evaluated using \gls{MF1} as the main metric.

\begin{table*}[!t]
\centering
\resizebox{\textwidth}{!}{
\begin{tabular}{@{}lccc|cccc@{}}
\toprule
\multirow{2}{*}{ID} & \multicolumn{3}{c|}{Model Configuration} & \multicolumn{4}{c}{\gls{BAH} sub-corpus} \\
\cmidrule(lr){2-4} \cmidrule(l){5-8}
 & Modality & Features & Classifier & Devel. / Valid. (MF1, \%) & Test (MF1, \%) & Average (MF1, \%) & Final test (MF1, \%) \\
\midrule
1 & Face & EmotionEfficientNetB0 + Statistical Features & MLP & 65.29 &60.05 &62.67& -- \\
2 & Scene & VideoMAE & Linear Layer & 61.71 & 62.21 & 61.96 & --\\
3 & Audio & EmotionWav2Vec2.0 + Mamba& Linear Layer & 67.20 & 70.87 & 69.03 &  --\\
4 & Text & TF-IDF & Logistic Regression & 68.30 & 67.75 & 68.03 &  --\\
5& Text & TF-IDF & CatBoost & 65.56 & 72.02 & 68.79 & -- \\
6 & Text & Fine-tuned EmotionTextClassifier & MLP & 69.28 & 70.72 & 70.00 & --\\
7 & Text & Fine-tuned EmotionDistilRoBERTa & MLP & 68.54 & 71.49 & 70.02 & --\\
\midrule
8 & Models IDs 1, 2, 3 and 4 & Multimodal Fusion Model & Linear Layer & 80.79 & 77.03 & 78.91 & -- \\
9 & Models IDs 1, 2, 3 and 5 & Multimodal Fusion Model & Linear Layer & 77.91 & 78.54 & 78.22 & -- \\
10 & Models IDs 1, 2, 3 and 6 & Multimodal Fusion Model & Linear Layer & 78.35 & 77.03 & 77.69 & -- \\
11 & Models IDs 1, 2, 3 and 7 & Multimodal Fusion Model & Linear Layer & 85.38 & 79.94 & 82.66 & 68.32 \\
12 & Models IDs 1, 2, 3 and 7 & Multimodal Fusion Model with Prototype Head & Linear Layer & 83.79 & 82.72 & 83.25 & 65.21 \\
\midrule
13 & Models IDs 1, 2, 3 and 7 & Ensemble of Five Multimodal Fusion Models & Linear Layer & 81.94 & 80.64 & 81.29 & 70.17 \\
14 & Models IDs 1, 2, 3 and 7 & Ensemble of Five Multimodal Fusion Models with Prototype Head & Linear Layer & 83.00 & 80.77 & 81.89 & 71.43 \\
\midrule
\multicolumn{8}{c}{Ablation Study} \\
\midrule
15 & Models IDs 1 and 3 & Multimodal Fusion Model & Linear Layer & 63.36 & 71.44 & 67.40 & -- \\
16 & Models IDs 1 and 7 & Multimodal Fusion Model & Linear Layer & 65.29 & 61.19 & 63.24 & -- \\
17 & Models IDs 1 and 2 & Multimodal Fusion Model & Linear Layer & 78.07 & 77.09 & 77.58 & -- \\
18 & Models IDs 3 and 7 & Multimodal Fusion Model & Linear Layer & 67.05 & 70.99 & 69.02 & -- \\
19 & Models IDs 2 and 3 & Multimodal Fusion Model & Linear Layer & 77.37 & 77.66 & 77.51 & -- \\
20 & Models IDs 2 and 7 & Multimodal Fusion Model & Linear Layer & 81.77 & 79.00 & 80.39 & -- \\
21 & Models IDs 2, 3 and 7 & Multimodal Fusion Model & Linear Layer & 79.89 & 77.63 & 78.76 & -- \\
22 & Models IDs 1, 2 and 7 & Multimodal Fusion Model & Linear Layer & 79.89 & 77.65 & 78.77 & -- \\
23 & Models IDs 1, 2 and 3 & Multimodal Fusion Model & Linear Layer & 76.10 & 79.15 & 77.62 & -- \\
24 & Models IDs 1, 3 and 7 & Multimodal Fusion Model & Linear Layer & 68.08 & 70.41 & 69.25 & -- \\
\bottomrule
\end{tabular}
}
\caption{Experimental results on the \gls{BAH} corpus for video-level \gls{A/H} recognition.}
\label{tab:Results}
\end{table*}

\subsection{Experimental Setup}

During the development of the unimodal models, several alternative configurations were evaluated. Scene modeling was performed using $16$-frame sequences resized to $224 \times 224$, trained for $15$ epochs with AdamW, a \gls{LR} of $2e\text{-}5$, weight decay of $1e\text{-}2$, batch size $4$, cosine annealing, and \gls{LS} of $0.1$. For the face-based visual modality, both statistical features with an \gls{MLP} and raw embeddings extracted by \gls{ViT}~\cite{vit2020} and \gls{CLIP}~\cite{radford2021learning} in combination with Transformer~\cite{vaswani2017attention} and Mamba~\cite{gu2024mamba} encoders were considered. The best configuration was based on statistical features with an \gls{MLP}. A grid search over the number of frames, hidden states, output features, \gls{LR}, and optimizer selected a setup with $30$ frames, $16$ hidden states, $256$ output features, a \gls{LR} of $1e\text{-}3$, and AdamW.

For the acoustic modality, embeddings from different EmotionWav2Vec2.0 layers and different temporal encoders, including Mamba and Transformer, were evaluated. The best result was obtained with layer $10$ and Mamba. In all primary runs, AdamW was used. The selected acoustic setup employed hidden size $256$, feed-forward size $512$, dropout $0.1$, mean pooling, Mamba state size $8$, convolution kernel size $4$, and expansion factor $2$. In the linguistic modality, \gls{TF-IDF} features were evaluated with vocabulary sizes from $100$ to $10{,}000$ and $n$-grams from unigrams to trigrams, with hyperparameter tuning for Logistic Regression and \gls{GBM} models. Transformer-based text models were also fine-tuned using partially frozen backbones and jointly trained \gls{MLP} heads. The number of hidden layers varied from $1$ to $3$, the hidden size from $64$ to $128$, and dropout from $0$ to $0.3$. AdamW and \gls{SGD} were explored, the \gls{LR} was searched in the range from $1e\text{-}5$ to $0.1$, the batch size was set to $16$, and training lasted from $3$ to $20$ epochs with early stopping.

Multimodal fusion was performed using embeddings extracted from the selected scene, face, audio, and text models. Both non-prototype and prototype-augmented variants were evaluated, and the final system was based on the prototype-augmented fusion model operating on embedding-level inputs. To reduce sensitivity to random initialization, model selection relied on a stability-oriented hyperparameter search with Optuna~\cite{akiba2019optuna} using five fixed random seeds: $42$, $2025$, $7777$, $12345$, and $31415$. Each candidate configuration was trained and evaluated five times, and the final score was computed as the average \gls{MF1} across these runs. The selected fusion model used a shared latent dimensionality of $128$, $6$ Transformer encoder layers, $4$ attention heads, a feed-forward expansion factor of $6$, no \texttt{[CLS]} token, and dropout of $0.45$. The prototype head used $16$ learnable prototypes per class with temperature $\tau = 0.3$. Training was performed with RMSprop, a \gls{LR} of $9.44e\text{-}5$, weight decay of $5.55e\text{-}4$, \gls{LS} of $0.02$, gradient clipping of $0.5$, and a cosine learning-rate scheduler. The prototype loss weight was set to $\lambda_{\mathrm{proto}} = 0.2$, while the diversity regularization term was disabled, i.e., $\lambda_{\mathrm{div}} = 0$. Final predictions were obtained by averaging the class probabilities of the $5$ seed-specific models.

\subsection{Experimental Results}

The experimental results are presented in Table~\ref{tab:Results}. Among the unimodal models, the best average \gls{MF1} was obtained by the fine-tuned EmotionDistilRoBERTa model (70.02\%), followed closely by the fine-tuned EmotionTextClassifier (70.00\%) and the acoustic model based on EmotionWav2Vec2.0 and Mamba (69.03\%). The \gls{TF-IDF}-based text models also showed competitive results, reaching 68.03\% with Logistic Regression and 68.79\% with CatBoost. In contrast, the face- and scene-based models achieved lower average scores of 62.67\% and 61.96\%, respectively.

All multimodal fusion models outperformed the unimodal baselines. Among the single fusion models, the best average result was achieved by the prototype-augmented fusion model based on the selected scene, face, audio, and text modalities (ID 12), with 83.25\%, while the corresponding fusion model without the prototype head (ID 11) reached 82.66\%. These results indicate that both multimodal integration and prototype-based classification are beneficial under the development and public test settings.

The final test results show a different trend. Among the submitted four-modality systems, the best performance was obtained by the ensemble of five prototype-augmented fusion models (ID 14), which achieved 71.43\%. The ensemble of five fusion models without the prototype head (ID 13) reached 70.17\%, while the single fusion models achieved 68.32\% (ID 11) and 65.21\% (ID 12). Thus, although the prototype-augmented single model achieved the highest average result, ensembling was essential for the strongest final test performance and improved robustness on the private evaluation split.

The ablation study further confirms the benefit of combining modalities. The strongest two-modality result was obtained by combining scene and text features (ID 20), with an average \gls{MF1} of 80.39\%. Among the three-modality settings, the best performance was achieved by combining face, scene, and text features (ID 22), with 78.77\%. Overall, the best results were obtained with four-modality fusion.




\section{Conclusions}

This paper presented a multimodal approach for video-level \gls{A/H} recognition on the \gls{BAH} corpus. The proposed approach combined scene, face, audio, and text modalities and consistently outperformed unimodal baselines. Among the unimodal models, the best average \gls{MF1} of 70.02\% was achieved by the fine-tuned EmotionDistilRoBERTa model, confirming the strong contribution of the linguistic modality. At the multimodal level, the best average result of 83.25\% was obtained by the prototype-augmented fusion model, while the best final test performance of 71.43\% was achieved by the ensemble of five prototype-augmented fusion models.

The ablation study showed that scene and text provide the strongest complementary signal among modality pairs, and that extending the fusion scheme to all four modalities yields the most effective overall solution. The comparison between single fusion models and their ensembles further indicates that robust model aggregation is important for generalization on the private test split. Overall, these results demonstrate that prototype-augmented multimodal fusion is an effective and robust strategy for \gls{A/H} recognition in unconstrained videos.

{
    \small
    \bibliographystyle{ieeenat_fullname}
    \bibliography{main}

@String(NATURE = {Nature})

@String(TAC = {IEEE Trans. Affect. Comput.})

@String(AEJ = {Alex. Eng. J.})

@String(NEUROCOM = {Neurocomputing})

@String(CSR = {Comput. Sci. Rev.})

@String(IF = {Inf. Fusion})

@String(JCBS = {J. Context. Behav. Sci.})

@String(NEURALCOMP = {Neural Comput.})

@String(TPAMI = {IEEE Trans. Pattern Anal. Mach. Intell.})

@String(CACM = {Commun. ACM})

@String(NIPS  = {NeurIPS})

@String(ICLR  = {ICLR})

@String(CVPR  = {CVPR})

@String(ICCV  = {ICCV})

@String(CVPRW = {CVPRW})

@String(NAACL = {NAACL})

@String(ICML  = {ICML})

@String(COLM  = {CoLM})

@String(KDD = {KDD})

@article{sajjad2023comprehensive,
    title={A comprehensive survey on deep facial expression recognition: challenges, applications, and future guidelines},
    author={Sajjad, Muhammad and Ullah, Fath U Min and Ullah, Mohib and Christodoulou, Georgia and Cheikh, Faouzi Alaya and Hijji, Mohammad and Muhammad, Khan and Rodrigues, Joel JPC},
    journal=AEJ,
    volume={68},
    pages={817--840},
    year={2023}
}

@article{george2024review,
    title={A review on speech emotion recognition: A survey, recent advances, challenges, and the influence of noise},
    author={George, Swapna Mol and Ilyas, P Muhamed},
    journal=NEUROCOM,
    volume={568},
    pages={1--23},
    year={2024}
}

@article{deng2021survey,
    title={A survey of textual emotion recognition and its challenges},
    author={Deng, Jiawen and Ren, Fuji},
    journal=TAC,
    volume={14},
    number={1},
    pages={49--67},
    year={2021}
}

@article{leong2023facial,
    title={Facial expression and body gesture emotion recognition: A systematic review on the use of visual data in affective computing},
    author={Leong, Sze Chit and Tang, Yuk Ming and Lai, Chung Hin and Lee, Carman Ka Man},
    journal=CSR,
    volume={48},
    pages={1--13},
    year={2023}
}

@article{poria2017review,
    title   = {A Review of Affective Computing: From Unimodal Analysis to Multimodal Fusion},
    author  = {Poria, Soujanya and Cambria, Erik and Bajpai, Rajiv and Hussain, Amir},
    journal = IF,
    volume  = {37},
    pages   = {98--125},
    year    = {2017}
}

@inproceedings{gonzalezbah,
    title={BAH Dataset for Ambivalence/Hesitancy Recognition in Videos for Digital Behavioural Change},
    author={Gonz{\'a}lez-Gonz{\'a}lez, Manuela and Belharbi, Soufiane and Zeeshan, Muhammad Osama and Sharafi, Masoumeh and Aslam, Muhammad Haseeb and Pedersoli, Marco and Koerich, Alessandro Lameiras and Bacon, Simon L and Granger, Eric},
    booktitle=ICLR,
    year={2026},
    pages = {1--17},
}

@article{bijkerk2024engagement,
    title={Engagement with mental health and health behavior change interventions: an integrative review of key concepts},
    author={Bijkerk, Laura E and Spigt, Mark and Oenema, Anke and Geschwind, Nicole},
    journal=JCBS,
    volume={32},
    pages={1--14},
    year={2024}
}

@InProceedings{Savchenko_2025_CVPR,
    author    = {Savchenko, Andrey and Savchenko, Lyudmila},
    title     = {Leveraging Lightweight Facial Models and Textual Modality in Audio-visual Emotional Understanding in-the-Wild},
    booktitle = CVPRW,
    year      = {2025},
    pages     = {5824--5834}
}

@InProceedings{Hallmen_2025_CVPR,
    author    = {Hallmen, Tobias and Kampa, Robin-Nico and Deuser, Fabian and Oswald, Norbert and Andr\'e, Elisabeth},
    title     = {Semantic Matters: Multimodal Features for Affective Analysis},
    booktitle = CVPRW,
    year      = {2025},
    pages     = {5761--5770}
}

@inproceedings{caron2021emerging,
    title={Emerging properties in self-supervised vision transformers},
    author={Caron, Mathilde and Touvron, Hugo and Misra, Ishan and J{\'e}gou, Herv{\'e} and Mairal, Julien and Bojanowski, Piotr and Joulin, Armand},
    booktitle=ICCV,
    pages={9650--9660},
    year={2021}
}

@inproceedings{baevski2020wav2vec,
    title={Wav2vec 2.0: A framework for self-supervised learning of speech representations},
    author={Baevski, Alexei and Zhou, Yuhao and Mohamed, Abdelrahman and Auli, Michael},
    booktitle=NIPS,
    pages={12449--12460},
    year={2020}
}

@inproceedings{devlin2019bert,
    title={Bert: Pre-training of deep bidirectional transformers for language understanding},
    author={Devlin, Jacob and Chang, Ming-Wei and Lee, Kenton and Toutanova, Kristina},
    booktitle=NAACL,
    pages={4171--4186},
    year={2019}
}

@article{hochreiter1997long,
    title={Long short-term memory},
    author={Hochreiter, Sepp and Schmidhuber, J{\"u}rgen},
    journal=NEURALCOMP,
    volume={9},
    number={8},
    pages={1735--1780},
    year={1997}
}

@article{rumelhart1986learning,
    title={Learning representations by back-propagating errors},
    author={Rumelhart, David E and Hinton, Geoffrey E and Williams, Ronald J},
    journal=NATURE,
    volume={323},
    number={6088},
    pages={533--536},
    year={1986}
}

@inproceedings{savchenko2024emotieffnet,
    title={EmotiEffNet and temporal convolutional networks in video-based facial expression recognition and action unit detection},
    author={Savchenko, Andrey V and Sidorova, Anna P},
    booktitle=CVPRW,
    pages={4849--4859},
    year={2024}
}

@article{liu2019roberta,
    title={Roberta: A robustly optimized bert pretraining approach},
    author={Liu, Yinhan and Ott, Myle and Goyal, Naman and Du, Jingfei and Joshi, Mandar and Chen, Danqi and Levy, Omer and Lewis, Mike and Zettlemoyer, Luke and Stoyanov, Veselin},
    journal={arXiv preprint arXiv:1907.11692},
    year={2019}
}

@inproceedings{tong2022videomae,
    title={VideoMAE: Masked Autoencoders are Data-Efficient Learners for Self-Supervised Video Pre-Training},
    author={Zhan Tong and Yibing Song and Jue Wang and Limin Wang},
    booktitle=NIPS,
    pages={1--15},
    year={2022}
}

@inproceedings{vit2020,
    title	= {An Image is Worth 16x16 Words: Transformers for Image Recognition at Scale},
    author	= {Dosovitskiy, Alexey and Beyer, Lucas and Kolesnikov, Alexander and Weissenborn, Dirk and Zhai, Xiaohua and Unterthiner, Thomas and Dehghani, Mostafa and Minderer, Matthias and Heigold, Georg and Gelly, Sylvain and others},
    booktitle = ICLR,
    pages = {1--22},
    year	= {2021}
}

@article{kay2017kinetics,
    title={The kinetics human action video dataset},
    author={Will Kay and Joao Carreira and Karen Simonyan and Brian Zhang and Chloe Hillier and Sudheendra Vijayanarasimhan and Fabio Viola and Tim Green and Trevor Back and Paul Natsev and Mustafa Suleyman and Andrew Zisserman},
    journal={arXiv preprint arXiv:1705.06950},
    year={2017},
    pages={1--22},
}

@inproceedings{deng2009imagenet,
    title={Imagenet: A large-scale hierarchical image database},
    author={Deng, Jia and Dong, Wei and Socher, Richard and Li, Li-Jia and Li, Kai and Fei-Fei, Li},
    booktitle = CVPR,
    pages={248--255},
    year={2009},
}

@inproceedings{tan2019efficientnet,
    title={EfficientNet: Rethinking model scaling for convolutional neural networks},
    author={Tan, Mingxing and Le, Quoc},
    booktitle=ICML,
    pages={6105--6114},
    year={2019},
}

@article{fard2025affectnetplus,
    title={Affectnet+: A database for enhancing facial expression recognition with soft-labels},
    author={Fard, Ali Pourramezan and Hosseini, Mohammad Mehdi and Sweeny, Timothy D and Mahoor, Mohammad H},
    journal=TAC,
    year={2026},
    volume={17},
    pages={784--800},
}

@article{Wagner_2023,
    title={Dawn of the Transformer Era in Speech Emotion Recognition: Closing the Valence Gap},
    journal=TPAMI,
    author={Wagner, Johannes and Triantafyllopoulos, Andreas and Wierstorf, Hagen and Schmitt, Maximilian and Burkhardt, Felix and Eyben, Florian and Schuller, Björn W.},
    year={2023},
    pages={10745--10759},
    doi = {10.1109/TPAMI.2023.3263585}
}

@article{lotfian2017building,
    title={Building naturalistic emotionally balanced speech corpus by retrieving emotional speech from existing podcast recordings},
    author={Lotfian, Reza and Busso, Carlos},
    journal=TAC,
    volume={10},
    number={4},
    pages={471--483},
    year={2017},
    doi={10.1109/TAFFC.2017.2736999}
}

@inproceedings{vaswani2017attention,
    title={Attention is all you need},
    author={Vaswani, Ashish and Shazeer, Noam and Parmar, Niki and Uszkoreit, Jakob and Jones, Llion and Gomez, Aidan N and Kaiser, {\L}ukasz and Polosukhin, Illia},
    booktitle=NIPS,
    year={2017},
    pages={1--11},
}

@inproceedings{gu2024mamba,
    title={Mamba: Linear-time sequence modeling with selective state spaces},
    author={Gu, Albert and Dao, Tri},
    booktitle=COLM,
    pages = {1--16},
    year	= {2024}
}

@article{saltonVectorSpaceModel1975,
    title = {A Vector Space Model for Automatic Indexing},
    author = {Salton, G. and Wong, A. and Yang, C. S.},
    year = 1975,
    month = nov,
    journal = CACM,
    volume = {18},
    number = {11},
    pages = {613--620},
    doi = {10.1145/361219.361220},
}

@inproceedings{keLightGBMHighlyEfficient2017,
    title = {LightGBM: A Highly Efficient Gradient Boosting Decision Tree},
    booktitle = NIPS,
    author = {Ke, Guolin and Meng, Qi and Finley, Thomas and Wang, Taifeng and Chen, Wei and Ma, Weidong and Ye, Qiwei and Liu, Tie-Yan},
    year = 2017,
    pages = {1--9},
}

@inproceedings{prokhorenkova2018catboost,
    title={CatBoost: unbiased boosting with categorical features},
    author={Prokhorenkova, Liudmila and Gusev, Gleb and Vorobev, Aleksandr and Dorogush, Anna Veronika and Gulin, Andrey},
    booktitle = NIPS,
    year={2018},
    pages = {1--11},
}

@inproceedings{he2021deberta,
    title={DeBERTa: Decoding-enhanced bert with disentangled attention},
    author={Pengcheng He and Xiaodong Liu and Jianfeng Gao and Weizhu Chen},
    booktitle=ICLR,
    year={2021},
    pages = {1--21},
}

@inproceedings{akiba2019optuna,
    title={Optuna: A next-generation hyperparameter optimization framework},
    author={Akiba, Takuya and Sano, Shotaro and Yanase, Toshihiko and Ohta, Takeru and Koyama, Masanori},
    booktitle= KDD,
    pages={2623--2631},
    year={2019}
}

@inproceedings{fang2025emoe,
    title={Emoe: Modality-specific enhanced dynamic emotion experts},
    author={Fang, Yiyang and Huang, Wenke and Wan, Guancheng and Su, Kehua and Ye, Mang},
    booktitle=CVPR,
    pages={14314--14324},
    year={2025}
}

@inproceedings{yang2025uncertain,
    title={Uncertain multimodal intention and emotion understanding in the wild},
    author={Yang, Qu and Shi, Qinghongya and Wang, Tongxin and Ye, Mang},
    booktitle=CVPR,
    pages={24700--24709},
    year={2025}
}

@inproceedings{li2025infobridge,
    title={InfoBridge: Balanced Multimodal Integration through Conditional Dependency Modeling},
    author={Li, Chenxin and Liu, Yifan and Pan, Panwang and Liu, Hengyu and Liu, Xinyu and Li, Wuyang and Wang, Cheng and Yu, Weihao and Lin, Yiyang and Yuan, Yixuan},
    booktitle=ICCV,
    pages={393--404},
    year={2025}
}

@inproceedings{radford2021learning,
    title={Learning transferable visual models from natural language supervision},
    author={Radford, Alec and Kim, Jong Wook and Hallacy, Chris and Ramesh, Aditya and Goh, Gabriel and Agarwal, Sandhini and Sastry, Girish and Askell, Amanda and Mishkin, Pamela and Clark, Jack and Krueger, Gretchen and Sutskever, Ilya},
    booktitle=ICML,
    pages={8748--8763},
    year={2021},
}
}


\end{document}